\documentclass[journal]{IEEEtran}
\usepackage{amsfonts}
\usepackage{amsmath}
\usepackage{amssymb}
\usepackage{cite}
\usepackage[colorlinks, linkcolor=red, anchorcolor=blue, citecolor=green]{hyperref}
\usepackage{bm}
\usepackage{graphicx}
\usepackage[dvipsnames]{xcolor}
\usepackage{booktabs}
\usepackage{subfigure}
\usepackage{threeparttable}
\usepackage{multirow}
\usepackage{rotating}
\usepackage{changepage}
\usepackage{makecell}
\usepackage{floatrow}
\usepackage{color, colortbl}
\usepackage{url}
\usepackage{pifont}
\usepackage{xcolor}

\newcommand{\RR}[2]{\textcolor[rgb]{0,0,0}
{#2}}

\newcommand{\cmark}{\textcolor{green!80!black}{\ding{51}}}
\newcommand{\xmark}{\textcolor{red}{\ding{55}}}

\definecolor{LightCyan}{rgb}{0.49, 0.78, 0.91}

\ifCLASSINFOpdf
\else
\fi

\begin{document}

\twocolumn[
  \begin{@twocolumnfalse}
    This work has been submitted to the IEEE for possible publication. Copyright may be transferred without notice, after which this version may no longer be accessible.
  \end{@twocolumnfalse}
]

\newpage

\title{ERA: A Dataset and Deep Learning Benchmark for Event Recognition in Aerial Videos}
\author{Lichao~Mou,
       ~Yuansheng~Hua,
       ~Pu~Jin, and
       ~Xiao~Xiang~Zhu,~\IEEEmembership{Senior Member,~IEEE}

\thanks{This work is jointly supported by the European Research Council (ERC) under the European Union's Horizon 2020 research and innovation programme (grant agreement No. [ERC-2016-StG-714087], Acronym: \textit{So2Sat}), by the Helmholtz Association
through the Framework of Helmholtz Artificial Intelligence Cooperation Unit (HAICU) - Local Unit ``Munich Unit @Aeronautics, Space and Transport (MASTr)'' and Helmholtz Excellent Professorship ``Data Science in Earth Observation - Big Data Fusion for Urban Research'', and by the German Federal Ministry of Education and Research (BMBF) in the framework of the international future AI lab ``AI4EO -- Artificial Intelligence for Earth Observation: Reasoning, Uncertainties, Ethics and Beyond''.}

\thanks{
L. Mou, Y. Hua, and X. X. Zhu are with the Remote Sensing Technology Institute (IMF), German Aerospace Center (DLR), Germany and with the Signal Processing in Earth Observation (SiPEO), Technical University of Munich (TUM), Germany (e-mails: lichao.mou@dlr.de; yuansheng.hua@dlr.de; xiaoxiang.zhu@dlr.de).}
\thanks{P. Jin is with the Technical University of Munich (TUM), Germany (e-mail: pu.jin@tum.de).}
       }

\maketitle

\begin{abstract}
\textcolor{blue}{This is the pre-acceptance version. To read the final version, please go to IEEE Geoscience and Remote Sensing Magazine on IEEE Xplore.} Along with the increasing use of unmanned aerial vehicles (UAVs), large volumes of aerial videos have been produced. It is unrealistic for humans to screen such big data and understand their contents. Hence methodological research on the automatic understanding of UAV videos is of paramount importance. In this paper, we introduce a novel problem of event recognition in unconstrained aerial videos in the remote sensing community and present a large-scale, human-annotated dataset, named ERA (Event Recognition in Aerial videos), consisting of 2,864 videos each with a label from 25 different classes corresponding to an event unfolding 5 seconds. \RR{}{All these videos are collected from YouTube.} The ERA dataset is designed to have a significant intra-class variation and inter-class similarity and captures dynamic events in various circumstances and at dramatically various scales. Moreover, to offer a benchmark for this task, we extensively validate existing deep networks. We expect that the ERA dataset will facilitate further progress in automatic aerial video comprehension. \RR{}{The dataset and trained models can be downloaded from \url{https://lcmou.github.io/ERA_Dataset/}.}
\end{abstract}


\IEEEpeerreviewmaketitle

\section{Introduction}
\label{sec:intro}
\IEEEPARstart{U}{nmanned} aerial vehicles (UAVs), a.k.a. drones, get a bad reputation in the media. Most people associate them with negative news, such as flight delays \RR{}{caused by} unauthorized drone activities and dangerous attack weapons. However, recent advances in the field of remote sensing and computer vision showcase that the future of UAVs will actually be shaped by a wide range of practical applications~\cite{xiang2018mini,bhardwaj2016uavs,dlinrs}. To name a few, in the aftermath of earthquakes and floods, UAVs can be exploited to estimate damage, deliver assistance, and locate victims. In addition to disaster relief, urban planners are capable of better understanding the environment of a city and implementing data-driven improvements by the use of UAVs. In precision agriculture, agricultural workers can make use of UAVs to collect data, automate redundant procedures, and generally maximize efficiency. In combination with geospatial information, UAVs are now used to monitor and track animals for the purpose of wildlife conservation.

\begin{figure}[t]
\centering
\subfigure[]{%
\includegraphics[width=4.0cm]{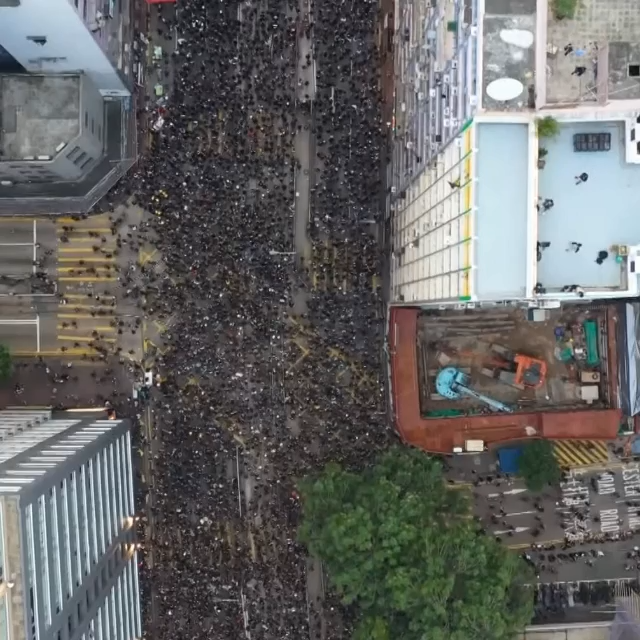}
\label{fig:subfigure1}}
\subfigure[]{%
\includegraphics[width=4.0cm]{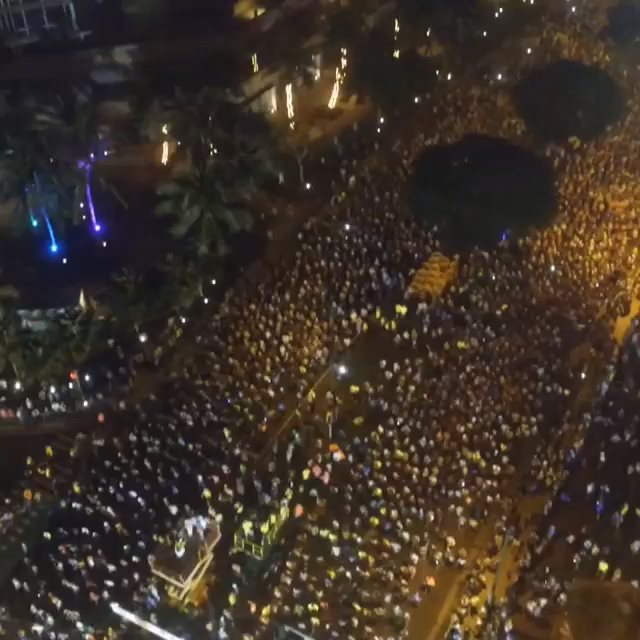}
\label{fig:subfigure2}}

\subfigure[]{%
\includegraphics[width=4.0cm]{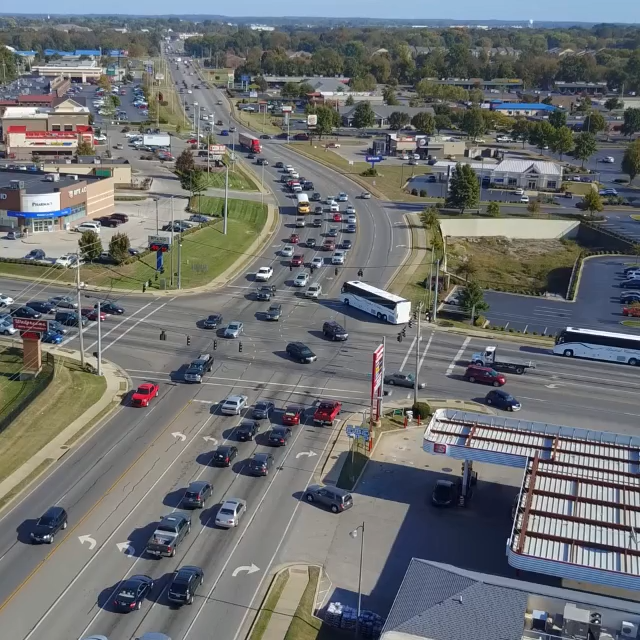}
\label{fig:subfigure3}}
\subfigure[]{%
\includegraphics[width=4.0cm]{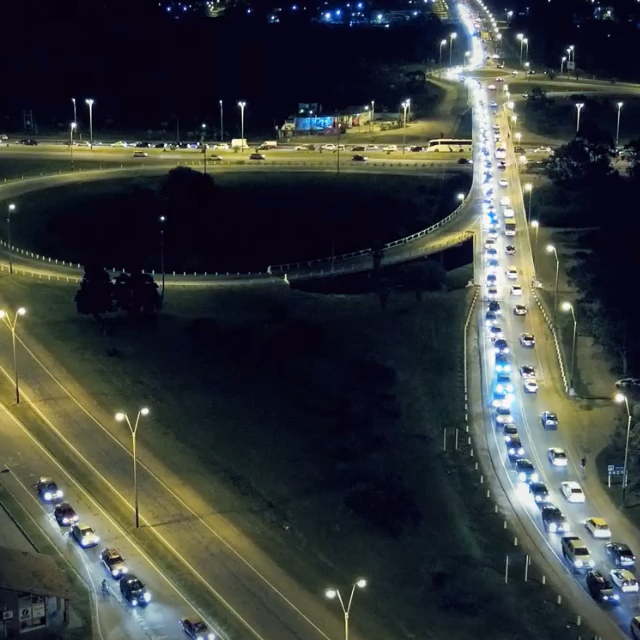}
\label{fig:subfigure4}}
\caption{\label{fig:temporal} \textbf{Temporal cue matters in event understanding from an aerial perspective.} What takes place in (a) and (b)? (c) or (d): Whose content depicts a traffic congestion scene? It is difficult to answer these questions from still images only, while temporal context provides an important visual cue. (See videos and answers at \url{https://lcmou.github.io/ERA_Dataset/gif_page/}.)}
\end{figure}

\begin{figure*}[t]
\centering
\includegraphics[width=0.95\textwidth]{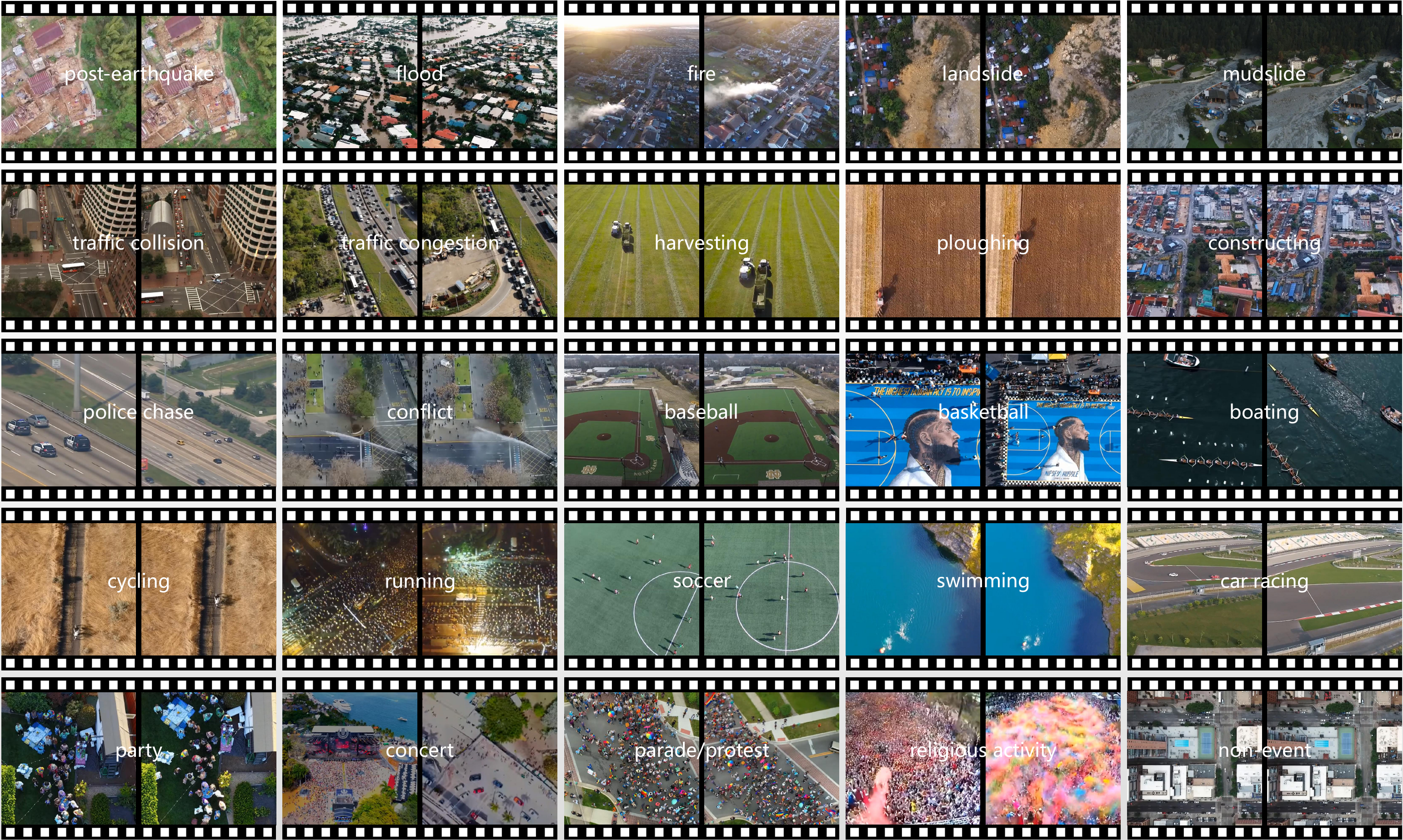}
\renewcommand{\figurename}{Fig.}
\caption{\label{fig:dataset} \textbf{Overview of the ERA dataset.} Overall, we have collected 2,864 labeled video snippets for 24 event classes and 1 normal class: \textit{post-earthquake}, \textit{flood}, \textit{fire}, \textit{landslide}, \textit{mudslide}, \textit{traffic collision}, \textit{traffic congestion}, \textit{harvesting}, \textit{ploughing}, \textit{constructing}, \textit{police chase}, \textit{conflict}, \textit{baseball}, \textit{basketball}, \textit{boating}, \textit{cycling}, \textit{running}, \textit{soccer}, \textit{swimming}, \textit{car racing}, \textit{party}, \textit{concert}, \textit{parade/protest}, \textit{religious activity}, and \textit{non-event}. For each class, we show the first (left) and last (right) frames of a video. Best
viewed zoomed in color.}
\end{figure*}

Unlike satellites, UAVs are able to provide real-time, high-resolution videos at a very low cost. They usually have real-time streaming capabilities that enable quick decision-making. Furthermore, UAVs can significantly reduce dependence on weather conditions, e.g., clouds, and are available on a demand offering higher flexibility to cope with various problems.
\par
Yet the more UAVs there are in the skies, the more video data they create. The Federal Aviation Administration (FAA) estimates that in the US alone, there are more than 2 million UAVs registered in 2019\footnote{\url{https://www.faa.gov/data_research/aviation/aerospace_forecasts/media/Unmanned_Aircraft_Systems.pdf}}. And every day around 150 terabytes of data can be easily produced by a small drone fleet\footnote{\url{https://www.bloomberg.com/news/articles/2017-05-10/airbus-joins-the-commercial-drone-data-wars}}. The era of big UAV data is here. It is unrealistic for humans to screen these massive volumes of aerial videos and understand their contents. Hence methodological research on the automatic interpretation of such data is of paramount importance.
\par
However, there is a paucity of literature on UAV video analysis, which for the most part is concentrated on detecting and tracking objects of interest~\cite{xiang2018mini,bhardwaj2016uavs,dlinrs}, e.g., vehicle and people, and understanding human activities in relatively sanitized settings~\cite{DBLP:conf/cvpr/ShuXRTZ15,DBLP:conf/cvpr/BarekatainMSMNM17}. Towards advancing aerial video parsing, this paper introduces a novel task, event recognition in unconstrained aerial videos, in the remote sensing community. We present an Event Recognition in Aerial video (ERA) dataset, a collection of 2,864 videos each with a label from 25 different classes corresponding to an event unfolding 5 seconds (see Fig.~\ref{fig:dataset}). \RR{}{Here each video is clipped from a YouTube long video, and its temporal length} (5 seconds) corresponds to the minimal duration of human short-term memory (5 to 20 seconds)~\cite{Smith75}. This dataset enables training models for richly understanding events in the wild from a broader, aerial view, which is a crucial step towards building an automatic aerial video comprehension system. In addition, to offer a benchmark for this task, we extensively validate existing deep networks and report their results in two ways: single-frame classification and video classification (see Section~\ref{sec:exp}).

\begin{figure}[t]
\centering
\includegraphics[width=0.94\columnwidth]{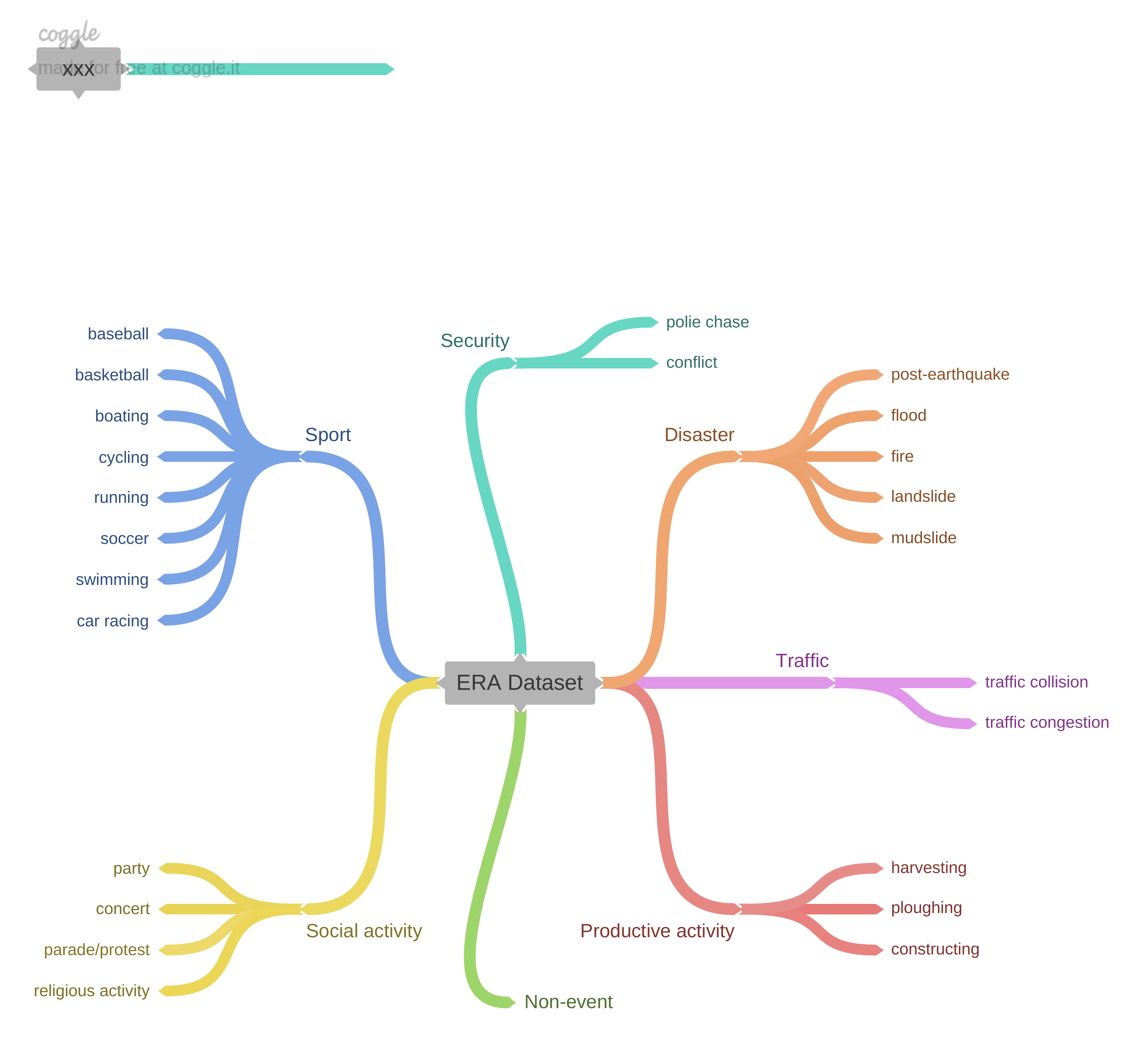}
\renewcommand{\figurename}{Fig.}
\caption{\label{fig:2} \textbf{Categorization of event classes in the ERA dataset.} All event categories are arranged in a two-level tree: with 25 leaf nodes connected to 7 nodes at the first level, i.e., \textit{disaster}, \textit{traffic}, \textit{productive activity}, \textit{security}, \textit{sport}, \textit{social activity}, and \textit{non-event}.}
\end{figure}

\begin{figure*}[t]
\centering
\includegraphics[width=0.95\textwidth]{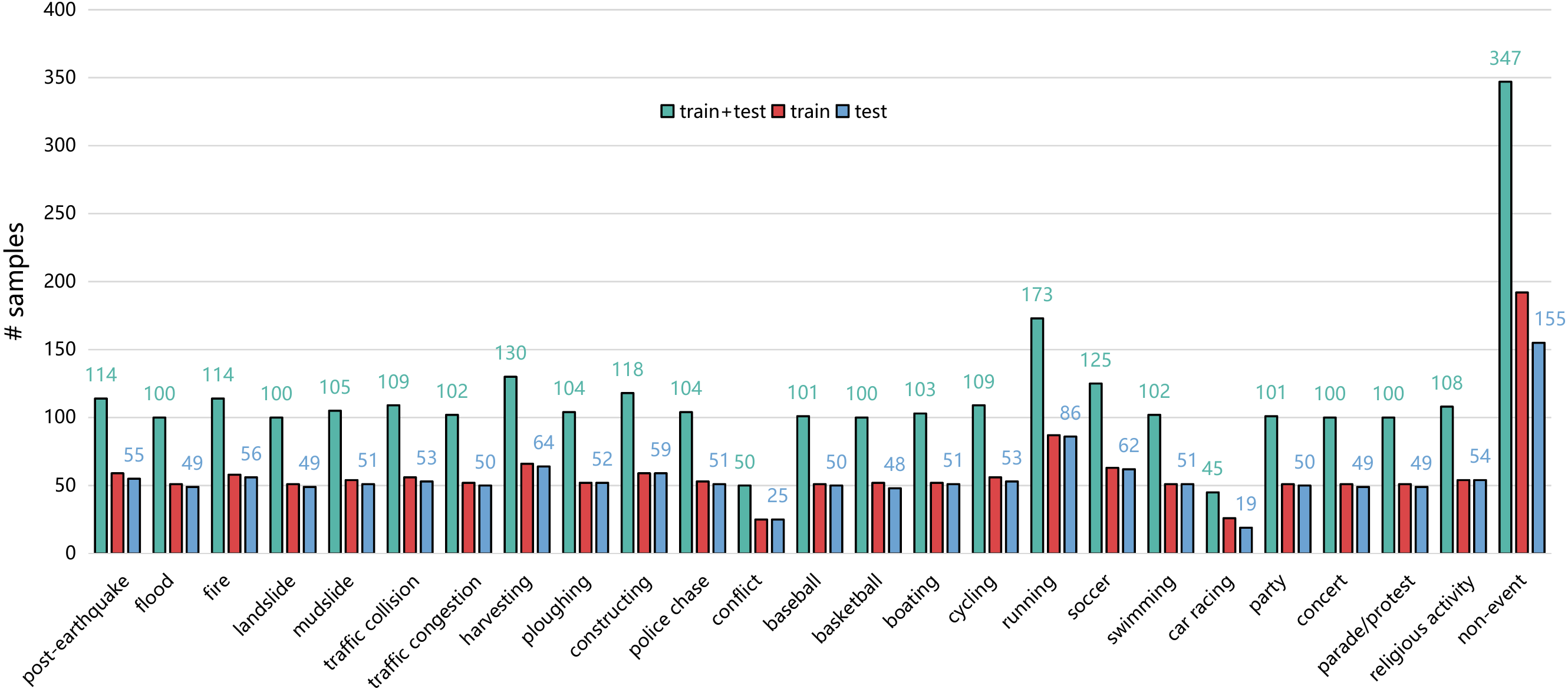}
\renewcommand{\figurename}{Fig.}
\caption{\label{fig:3} \textbf{Sample distributions of all classes in the ERA.} The red and blue bars represent the numbers of training and test samples in each category, respectively, and green bars denote the total number of instances in each category.}
\end{figure*}

\section{The ERA Dataset}
This work aims to devise an aerial video dataset which covers an extensive range of events. The ERA dataset is designed to have a significant intra-class variation and inter-class similarity and captures dynamic events in various circumstances and at dramatically various scales.

\subsection{Collection and Annotation}
We start by creating our taxonomy (cf. Fig.~\ref{fig:2}) by building a set of the 24 most commonly seen events from aerial scenes. Moreover, to investigate if models can distinguish events from normal videos, we set up a category called \textit{non-event}, which comprises videos not including specific events. The 25 classes of our dataset can be found in Fig.~\ref{fig:dataset}.
\par
In order to collect candidate videos for further labeling, we search YouTube by parsing the metadata of videos and crawling the search engine to create a collection of candidate videos for each category. \RR{}{As a result, we collect 1,120 long videos.} \RR{}{Note that we set \textit{drone} and \textit{UAV} as keywords in order to exclude videos not taken from UAVs.} Then we download all videos and send them to data annotators. Each annotator is asked to localize 5-second video snippets that depict specific events and cut them from long candidate videos.
\par
There are three different annotators with payments who are responsible for validating cut videos during the annotation procedure. The first annotator generates primary annotations to begin with. Then annotations from the first round are sent to the second annotator for tune-ups. Finally, the third annotator screen all generated 5-second videos and remove videos with quite similar contents. \RR{}{Moreover, they are asked to double-check whether these videos are taken by UAVs.} Overall, the total annotation time is around 290 hours.

\subsection{Dataset Statistics}
The goal of this work is to collect a large, diverse dataset that can be used to train models for event recognition in UAV videos. As we gather aerial videos from YouTube, the largest video sharing platform in the world, we are capable of including a large breadth of diversity, which is more challenging than making use of self-collected data~\cite{DBLP:conf/cvpr/ShuXRTZ15,DBLP:conf/cvpr/BarekatainMSMNM17}. In total, we have gathered and annotated 2,864 videos for 25 classes. Each video sequence is at 24 fps (frames per second), in 5 seconds, and with a spatial size of 640$\times$640 pixels. The train/test split can be found in Fig.~\ref{fig:3} and Section~\ref{sec:exp_setup}. Fig.~\ref{fig:3} exhibits the distribution of all classes. The red and blue bars represent the numbers of training and test samples in each category, respectively, and green bars denote the total number of instances in each category.
\par
To build a diverse dataset, we collect not only high quality UAV videos but also ones acquired in extreme conditions. By doing so, many challenging issues of event recognition in overhead videos in real-world scenarios, e.g., low spatial resolution, extreme illumination conditions, and bad weathers, can be investigated. True aerial video parsing methods should be capable of recognizing events under such extreme conditions.

\begin{table*}[t!]
\footnotesize
\caption{\textbf{Comparison to existing UAV data understanding datasets.} We offer various comparisons for each dataset.}
\label{tab:comparison}
\centering
\begin{tabular}{r|cccccc}
\Xhline{2\arrayrulewidth}
\textbf{Dataset} & \textbf{Type of Task} & \textbf{Data Source} & \textbf{Video} & \textbf{\# Classes} & \textbf{\# Samples} & \textbf{Year} \\
\hline
\hline
UCLA Aerial Event dataset~\cite{DBLP:conf/cvpr/ShuXRTZ15} & human-centric event recognition & self-collected (actor staged) & \cmark & 12 & 104 & 2015 \\
Okutama-Action dataset~\cite{DBLP:conf/cvpr/BarekatainMSMNM17} & human action detection & self-collected (actor staged) & \cmark & 12 & - & 2017 \\
AIDER dataset~\cite{DBLP:conf/cvpr/KyrkouT19} & disaster event recognition & Internet & \xmark & 5 & 2,545 & 2019 \\
\textbf{ERA dataset (Ours)} & \textbf{general event recognition} & \textbf{YouTube} & \cmark & \textbf{25} & \textbf{2,864} & \textbf{2020} \\
\Xhline{2\arrayrulewidth}
\end{tabular}
\end{table*}

\subsection{Comparison with Other Aerial Data Understanding Datasets}
The first significant effort to build a standard dataset for aerial video content understanding can be found in~\cite{DBLP:conf/cvpr/ShuXRTZ15}, in which the authors make use of a GoPro-equipped drone to collect video data at an altitude of 25 meters in a controlled environment to build a dataset called UCLA Aerial Event dataset. There are about 15 actors involved in each video. This dataset includes two different sites at a park in Los Angeles, USA and 104 event instances that present 12 classes related to human-human and human-object interactions.
\par
The authors of~\cite{DBLP:conf/cvpr/BarekatainMSMNM17} propose Okutama-Action dataset for understanding human actions from a bird's eye view. Two UAVs (DJI Phantom 4) with a flying altitude of 10-45 meters above the ground are used to capture data, and all videos included in this dataset are gathered at a baseball field in Okutama, Japan. There are 12 actions included.

In~\cite{DBLP:conf/cvpr/KyrkouT19}, the authors build an aerial image dataset, termed as AIDER, for emergency response applications. This dataset only involves four disaster events, namely \textit{fire/smoke}, \textit{flood}, \textit{collapsed building/rubble}, and \textit{traffic accident}, and a \textit{normal} case. There are totally 2,545 images collected from multiple sources, e.g., Google/Bing Images, websites of news agencies, and YouTube.
\par
Both the UCLA Aerial Event dataset~\cite{DBLP:conf/cvpr/ShuXRTZ15} and the Okutama-Action dataset~\cite{DBLP:conf/cvpr/BarekatainMSMNM17} are small in today’s terms for aerial video understanding, and their data are gathered in well-controlled environments and only focus on several human-centric events. \RR{}{Besides, the AIDER dataset}~\cite{DBLP:conf/cvpr/KyrkouT19} is an image dataset with only 5 classes for disaster event classification. In contrast, our ERA is a relatively large-scale UAV video content understanding dataset, aiming to recognize generic dynamic events from an aerial view. A comprehensive overview of these most important comparable datasets and their features is given in Table~\ref{tab:comparison}.

\subsection{Challenges}
The proposed ERA dataset poses the following challenges:

\begin{itemize}
  \item \RR{}{Although, to the best of our knowledge, the ERA dataset is the largest dataset for event recognition in aerial videos yet,} its size is still relatively limited as compared to video classification datasets in computer vision. Hence there exists the small data challenge in the model training.
  \item The imbalanced distribution across different classes (cf.~Fig.~\ref{fig:3}) brings a challenge of learning unbiased models on an imbalanced dataset.
  \item Unlike~\cite{DBLP:conf/cvpr/ShuXRTZ15,DBLP:conf/cvpr/BarekatainMSMNM17}, the existence of the \textit{non-event} class in our dataset requires that models are able to not only recognize different events but also distinguish events from normal videos.
  \item In this dataset, events happen in various environments and are observed at different scales, which leads to a significant intra-class variation and inter-class similarity.
\end{itemize}

\begin{table*}[t!]
\footnotesize
\caption{\textbf{Performance of single-frame classification models:} We show the per-class precision and overall accuracy of baseline models on the test set. The best precision/accuracy is shown in bold.}
\label{tab:acc_mid_frame}
\centering
\begin{adjustwidth}{-0.2cm}{0cm}
\begin{threeparttable}
\begin{tabular}{p{1.75cm}|*{25}{p{0.2cm}<{\centering}}p{0.4cm}}
\Xhline{2\arrayrulewidth}
\textbf{Model} & \rotatebox{75}{post-earthquake} & \rotatebox{75}{flood} & \rotatebox{75}{fire} & \rotatebox{75}{landslide} & \rotatebox{75}{mudslide} & \rotatebox{75}{traffic collision} & \rotatebox{75}{traffic congestion} & \rotatebox{75}{harvesting} & \rotatebox{75}{ploughing} & \rotatebox{75}{constructing} & \rotatebox{75}{police chase} & \rotatebox{75}{conflict} & \rotatebox{75}{baseball} & \rotatebox{75}{basketball} & \rotatebox{75}{boating} & \rotatebox{75}{cycling} & \rotatebox{75}{running} & \rotatebox{75}{soccer} & \rotatebox{75}{swimming} & \rotatebox{75}{car racing} & \rotatebox{75}{party} & \rotatebox{75}{concert} & \rotatebox{75}{parade/protest} & \rotatebox{75}{religious activity} & \rotatebox{75}{non-event} & \textbf{OA} \\
\hline
\hline
VGG-16 & 46.3 & 59.1 & 53.6 & 38.8 & 56.1 & 30.8 & \textbf{76.2} & 62.7 & 65.4 & 69.0 & 70.0 & 44.4 & 61.0 & 56.2 & 69.4 & 39.6 & 33.3 & \textbf{87.5} & 62.0 & 32.0 & \textbf{73.5} & 56.7 & 47.8 & 64.6 & 30.4 & 51.9 \\
VGG-19 & 45.5 & 56.4 & 70.2 & 48.1 & 47.1 & 33.3 & 50.0 & 57.1 & 58.1 & 65.2 & \textbf{80.9} & 7.4 & 66.7 & 55.9 & 66.7 & 35.8 & 57.1 & 67.3 & 55.6 & 26.7 & 53.3 & 54.4 & 43.6 & 50.0 & 31.1 & 49.7 \\
Inception-v3 & 62.9 & 76.1 & \textbf{88.0} & 44.7 & 54.7 & \textbf{48.0} & 55.4 & 64.6 & 77.3 & 73.7 & 76.5 & 50.0 & 72.0 & 61.2 & 73.7 & 70.2 & \textbf{90.0} & 80.0 & 61.7 & 60.0 & 66.7 & 47.7 & 52.2 & 62.2 & 45.5 & 62.1 \\
ResNet-50 & 65.5 & 69.8 & 77.4 & 40.0 & 51.9 & 40.6 & 50.0 & \textbf{77.4} & 72.9 & 63.8 & 68.6 & \textbf{62.5} & 83.3 & 52.2 & 71.4 & 77.4 & 28.6 & 73.5 & 54.3 & 50.0 & 61.5 & 49.4 & 46.0 & 48.9 & 38.9 & 57.3 \\
ResNet-101 & 59.6 & 82.9 & 79.2 & 34.5 & 43.8 & 18.8 & 48.7 & 65.8 & 78.0 & 69.5 & 64.6 & 55.0 & 76.1 & 57.7 & 82.2 & \textbf{90.5} & 61.5 & 73.3 & 58.2 & 31.6 & 51.2 & 49.5 & 47.1 & \textbf{64.7} & 36.2 & 55.3 \\
ResNet-152 & 67.3 & 68.2 & 78.8 & 45.2 & 46.4 & 38.9 & 58.5 & 61.9 & 75.6 & 58.0 & 59.3 & 57.1 & 79.5 & 56.9 & 77.8 & 63.4 & 75.0 & 74.4 & 56.1 & 30.8 & 61.9 & 44.7 & 48.6 & 52.8 & 37.0 & 56.1 \\
MobileNet & \textbf{72.0} & 70.8 & 78.0 & \textbf{57.5} & \textbf{61.0} & 43.6 & 52.6 & 66.2 & 66.7 & 67.2 & 70.6 & 50.0 & 74.5 & 59.7 & 76.4 & 54.7 & 72.0 & 64.8 & 52.9 & 56.2 & 65.0 & 44.4 & 54.5 & 61.5 & \textbf{52.5} & 61.3 \\
DenseNet-121 & 58.6 & 71.4 & 82.8 & 54.5 & 51.6 & 38.1 & 58.2 & 71.1 & 78.0 & 70.2 & 73.5 & 48.0 & 85.0 & 68.4 & \textbf{86.7} & 65.3 & 57.1 & 75.4 & 61.7 & 52.9 & 68.3 & 52.3 & \textbf{66.7} & 47.8 & 43.3 & 61.7 \\
DenseNet-169 & 70.0 & \textbf{82.9} & 71.9 & 45.2 & 40.2 & 36.7 & 59.5 & 71.6 & \textbf{87.2} & 80.4 & 76.6 & 53.8 & \textbf{91.4} & 65.0 & 67.7 & 76.9 & 63.6 & 75.0 & \textbf{63.2} & 57.1 & 59.1 & 60.0 & 55.4 & 60.9 & 39.7 & 60.6 \\
DenseNet-201 & 69.9 & 80.4 & 84.5 & 52.2 & 48.1 & 43.2 & 62.3 & 71.6 & 85.4 & 71.2 & 77.1 & 47.1 & 87.8 & 63.6 & 79.6 & 69.8 & 47.8 & 65.0 & 58.0 & 43.8 & 61.0 & \textbf{60.9} & 55.0 & 60.8 & 42.1 & \textbf{62.3} \\
NASNet-L & 60.0 & 50.0 & 77.2 & 41.0 & 50.9 & 46.9 & 50.0 & 68.0 & 77.8 & \textbf{82.7} & 78.0 & 61.5 & 82.6 & \textbf{74.5} & 78.0 & 75.0 & 62.2 & 69.0 & 54.5 & \textbf{70.0} & 69.2 & 44.6 & 58.7 & 55.9 & 41.7 & 60.2 \\
\Xhline{2\arrayrulewidth}
\end{tabular}

\begin{tablenotes}
    \item[1] All networks are initialized with weights pre-trained on the ImageNet dataset and trained on the ERA dataset.
\end{tablenotes}

\end{threeparttable}
\end{adjustwidth}
\end{table*}

\begin{table*}[t!]
\footnotesize
\caption{\textbf{Performance of video classification models:} We show the per-class precision and overall accuracy of baseline models on the test set. The best precision/accuracy is shown in bold.}
\label{tab:acc_video}
\centering
\begin{adjustwidth}{-0.2cm}{0cm}
\begin{threeparttable}
\begin{tabular}{p{1.75cm}|*{25}{p{0.2cm}<{\centering}}p{0.4cm}}
\Xhline{2\arrayrulewidth}
\textbf{Model} & \rotatebox{75}{post-earthquake} & \rotatebox{75}{flood} & \rotatebox{75}{fire} & \rotatebox{75}{landslide} & \rotatebox{75}{mudslide} & \rotatebox{75}{traffic collision} & \rotatebox{75}{traffic congestion} & \rotatebox{75}{harvesting} & \rotatebox{75}{ploughing} & \rotatebox{75}{constructing} & \rotatebox{75}{police chase} & \rotatebox{75}{conflict} & \rotatebox{75}{baseball} & \rotatebox{75}{basketball} & \rotatebox{75}{boating} & \rotatebox{75}{cycling} & \rotatebox{75}{running} & \rotatebox{75}{soccer} & \rotatebox{75}{swimming} & \rotatebox{75}{car racing} & \rotatebox{75}{party} & \rotatebox{75}{concert} & \rotatebox{75}{parade/protest} & \rotatebox{75}{religious activity} & \rotatebox{75}{non-event} & \textbf{OA} \\
\hline
\hline
C3D$^\dag$ & 23.1 & 24.3 & 30.9 & 19.5 & 32.9 & 7.00 & 15.5 & 27.5 & 36.1 & 45.5 & 50.0 & 18.2& 40.9 & 37.0 & 47.5 & 20.6 & 12.0 & 58.3 & 36.2 & 16.7 & 25.8 & 38.2 & 37.8 & 27.5 & 29.6 & 30.4 \\
C3D$^\ddag$ & 27.9 & 56.5 & 32.7 & 10.2 & 23.9 & 8.30 & 38.5 & 42.3 & 31.1 & 40.0 & 51.9 & 11.1 & 45.7 & 48.9 & 41.9 & 13.6 & 9.30 & 41.9 & 38.2 & 18.2 & 17.4 & 32.0 & 28.1 & 35.8 & 28.5 & 31.1 \\
P3D$^\dag$-{\tiny ResNet-199} & 43.6 & 65.9 & 66.7 & 35.5 & 48.7 & 20.0 & 37.8 & 77.4 & 70.8 & 62.0 & \textbf{81.6} & 22.2& 66.7 & 63.1 & 55.4 & 35.6 & 35.3 & 76.2 & 57.4 & 40.0 & 54.5 & 37.5 & 38.7 & 47.8 & 37.4 & 50.7 \\
P3D$^\ddag$-{\tiny ResNet-199} & 72.4 & 76.3 & 84.8 & 24.5 & 38.2 & 35.6 & 40.8 & 56.9 & 67.4 & 71.4 & 57.9 & 50.0 & 70.4 & 78.8 & 71.7 & 47.1 & 60.0 & 79.5 & 68.1 & 40.9 & 59.1 & 37.0 & 49.1 & 55.9 & 37.9 & 53.3 \\
I3D$^\dag$-{\tiny Inception-v1} & 40.4 & 63.5 & 68.9 & 22.6 & 46.3 & 17.6 & 55.0 & 61.5 & 50.0 & 53.3 & 73.2 & 50.0 & 75.0 & 69.4 & 60.7 & 61.9 & 53.3 & 70.8 & 52.5 & 50.0 & 57.1 & \textbf{50.7} & 40.3 & 49.0 & 35.8 & 51.3 \\
I3D$^\ddag$-{\tiny Inception-v1} & 60.0 & 68.1 & 65.7 & 29.0 & \textbf{60.4} & \textbf{51.5} & 52.2 & 67.1 & 66.7 & 54.2 & 64.8 & 57.9 & \textbf{85.0} & 61.9 & \textbf{86.4} & \textbf{75.0} & 44.4 & 77.6 & 64.1 & \textbf{65.2} & 53.7 & 50.0 & \textbf{47.8} & 65.1 & 43.0 & 58.5 \\
TRN$^\dag$-{\tiny BNInception} & \textbf{84.8} & 71.4 & 82.5 & 51.2 & 50.0 & 46.8 & \textbf{66.7} & 68.1 & 77.4 & 52.4 & 70.5 & \textbf{75.0} & 64.5 & 67.7 & 84.0 & 56.1 & 55.2 & \textbf{83.3} & \textbf{72.9} & 61.1 & 62.0 & 48.9 & 44.6 & 62.8 & \textbf{51.1} & 62.0 \\
TRN$^\ddag$-{\tiny Inception-v3} & 69.2 & \textbf{87.8} & \textbf{88.9} & \textbf{65.8} & 60.0 & 44.1 & 58.3 & \textbf{78.1} & \textbf{90.7} & \textbf{70.8} & 73.3 & 28.6 & 83.3 & \textbf{72.7} & 73.7 & 60.0 & \textbf{66.7} & 73.6 & 70.6 & 63.6 & \textbf{65.1} & 47.7 & 42.7 & \textbf{65.1} & 47.9 & \textbf{64.3} \\
\Xhline{2\arrayrulewidth}
\end{tabular}

\end{threeparttable}
\end{adjustwidth}
\end{table*}

\begin{figure*}[!t]
\centering
\includegraphics[width=0.95\textwidth]{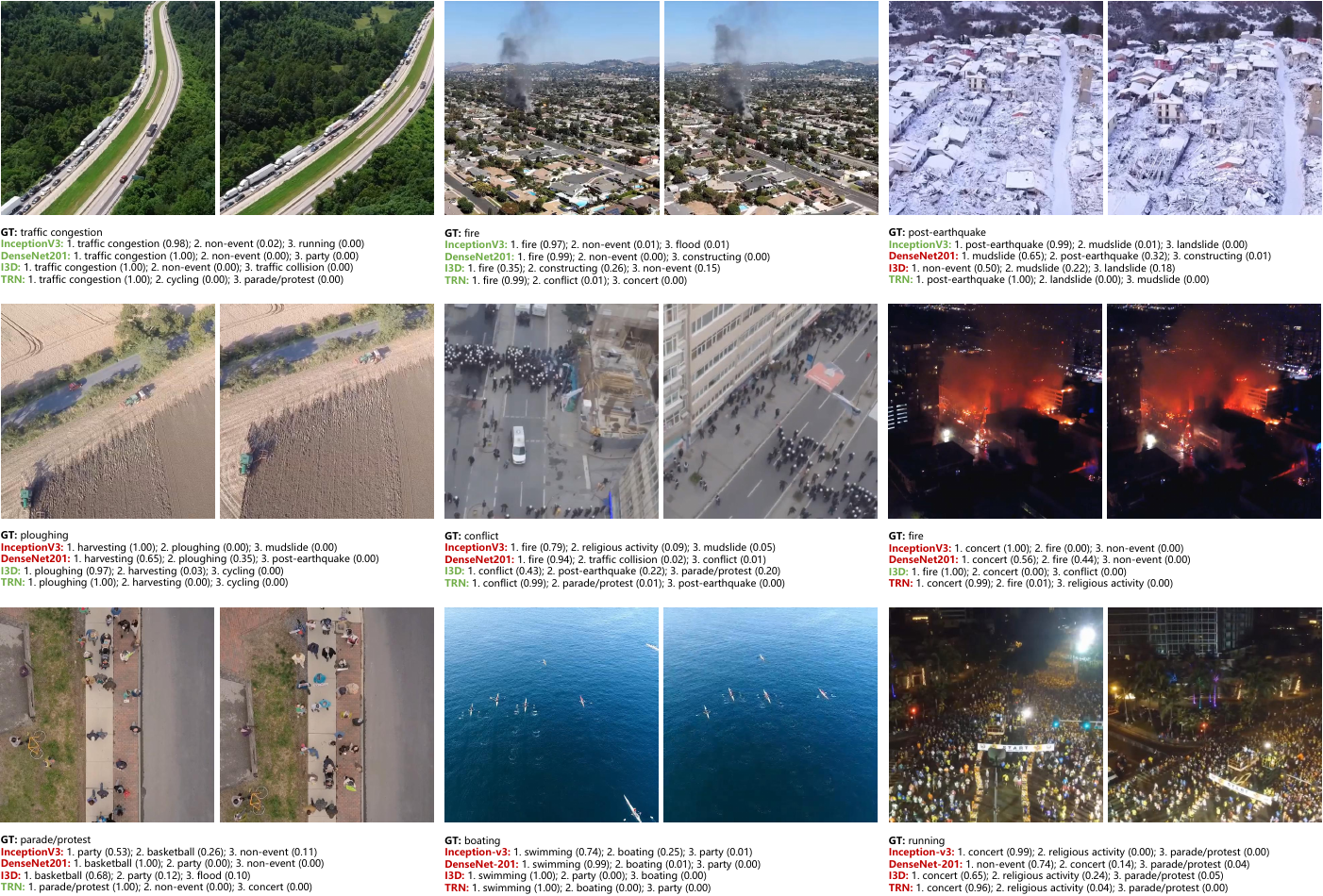}
\renewcommand{\figurename}{Fig.}
\caption{\label{fig:results} {\textbf{Examples of event recognition results on the ERA dataset.} We show the best two single-frame classification network architectures (i.e., Inception-v3 and DenseNet-201) and the best two video classification network architectures (i.e., I3D$^\ddag$-{\tiny Inception-v1} and TRN$^\ddag$-{\tiny Inception-v3}). The ground truth label and top 3 predictions of each model are reported. For each example, we show the first (left) and last (right) frames. Best viewed zoomed in color.} }
\end{figure*}

\section{Experiments}
\label{sec:exp}

\subsection{Experimental Setup}
\label{sec:exp_setup}
\textbf{Data and evaluation metric.} As to the split of training and test sets, we obey the following two rules: 1) videos cut from the same long video are assigned to the same set, and 2) the numbers of training and test videos per class are supposed to be nearly equivalent. Because video snippets stemming from the same long video usually share similar properties (e.g., background, illumination, and resolution), this split strategy is able to evaluate the generalization ability of a model. \RR{}{We have provided our training/test split in our dataset.}. The statistics of training and test samples are exhibited in Fig.~\ref{fig:3}. During the training phase, 10\% of training instances are randomly selected as the validation set. To compare models comprehensively, we make use of per-class precision and overall accuracy as evaluation metrics.

\par

\subsection{Baselines for Event Classification}

\textbf{Single-frame classification models.} We first describe single-frame classification models where only a single video frame is selected (the middle frame in this paper) from a video as the input to networks. The used single-frame models\footnote{\url{https://github.com/keras-team/keras-applications}} are as follows: VGG-16~\cite{vgg}, VGG19~\cite{vgg}, Inception-v3~\cite{inceptionv3}, ResNet-50~\cite{resnet}, ResNet-101~\cite{resnet}, ResNet-152~\cite{resnet}, MobileNet~\cite{mobilenet}, DenseNet-121~\cite{densenet}, DenseNet-169~\cite{densenet}, DenseNet-201~\cite{densenet}, and NASNet-L~\cite{nasnet}.


\textbf{Video classification models.} These models take several video frames as input, so that they can learn temporal information from videos. We summarize the used video classification models as follows.

\begin{itemize}
  \item C3D~\cite{tran2015learning}. In our experiments, we train two C3D\footnote{\url{https://github.com/tqvinhcs/C3D-tensorflow}} networks with pre-trained weights on the Sport1M dataset and the UCF101 dataset (see C3D$^\dag$ and C3D$^\ddag$ in Table~\ref{tab:acc_video}), respectively.
  \item P3D ResNet~\cite{p3d}. We train two 199-layer P3D ResNet\footnote{\url{https://github.com/zzy123abc/p3d}} (P3D-{\tiny ResNet-199}) models with pre-trained weights on the Kinetics dataset and the Kinetics-600 dataset (see P3D$^\dag$-{\tiny ResNet-199} and P3D$^\ddag$-{\tiny ResNet-199} in Table~\ref{tab:acc_video}), respectively.
  \item I3D~\cite{i3d}. To assess the performance of I3D on our dataset, we train two I3D\footnote{\url{https://github.com/LossNAN/I3D-Tensorflow}} models whose backbones are both Inception-v1~\cite{inception} (I3D-{\tiny Inception-v1}) with pre-trained weights on the Kinetics dataset and Kinetics+ImageNet, respectively (see I3D$^\dag$-{\tiny Inception-v1} and I3D$^\ddag$-{\tiny Inception-v1} in Table~\ref{tab:acc_video}).
  \item TRN~\cite{trn}. In our experiments, we train TRNs\footnote{\url{https://github.com/metalbubble/TRN-pytorch}} with 16 multi-scale relations and select the Inception architecture as the backbone. Notably, we experiment two variants of the Inception architecture: BNInception~\cite{DBLP:conf/icml/IoffeS15} and Inception-v3~\cite{inceptionv3}. We initialize the former with weights pre-trained on the Something-Something V2 dataset (TRN$^\dag$-{\tiny BNInception} in Table~\ref{tab:acc_video}) and the latter with weights pre-trained on the Moments in Time dataset (TRN$^\ddag$-{\tiny Inception-v3} in Table~\ref{tab:acc_video}).

\end{itemize}

\subsection{Baseline Results}
Quantitative results of single-frame classification models and video classification models are reported in Table~\ref{tab:acc_mid_frame} and Table~\ref{tab:acc_video}, respectively. As we can see, DenseNet-201 achieves the best performance, an OA of 62.3\%, in the single-frame classification task and marginally surpasses the second best model, Inception-v3, by 0.2\%. For the video classification task, TRN$^\ddag$-{\tiny Inception-v3} performs superiorly and gains an OA of 64.3\%. By comparing Table~\ref{tab:acc_mid_frame} and Table~\ref{tab:acc_video}, it is interesting to observe that the best-performed video classification model obtains the highest OA, which demonstrates the significance of exploiting temporal cues in event recognition from aerial videos.
\par
We further show some predictions of the best two single-frame classification network architectures (i.e., Inception-v3 and DenseNet-201) and the best two video classification network architectures (i.e., I3D$^\ddag$-{\tiny Inception-v1} and TRN$^\ddag$-{\tiny Inception-v3}) in Fig.~\ref{fig:results}. As shown in the top left two examples, frames/videos with discriminative event-relevant characteristics, such as congested traffic states on a highway and smoke rising from a residential area, can be accurately recognized by all baselines with high confidence scores. Besides, high-scoring predictions of TRN in identifying ploughing and parade/protest illustrate that efficiently exploiting temporal information helps in distinguishing events of minor inter-class variances. Moreover, we observe that extreme conditions might disturb predictions, for instance, frames/videos of night and snow scenes (see Fig.~\ref{fig:results}) tend to be misclassified.
\par
Despite successes achieved by these baselines, there are still some challenging cases as shown in Fig.~\ref{fig:results}. A common characteristic shared by these examples is that event-relevant attributes such as human actions are not easy to recognize, and this results in failures to identify these events. To summarize, event recognition in aerial videos is still a big challenge and may benefit from better recognizing discriminative attributes as well as exploiting temporal cues. More examples are at \url{https://lcmou.github.io/ERA_Dataset/}.

\section{Conclusion}
We present ERA, a dataset for comprehensively recognizing events in the wild form UAV videos. Organized in a rich semantic taxonomy, the ERA dataset covers a wide range of events involving diverse environments and scales. We report results of plenty of deep networks in two ways: single-frame classification and video classification. The experimental results show that this is a hard task for the remote sensing field, and the proposed dataset serves as a new challenge to develop models that can understand what happens on the planet from an aerial view. \RR{}{Looking into the future, our dataset has the potential of being applied to more tasks with existing or new annotations, e.g., temporal event localization in long videos, multi-attribute learning for aerial video understanding, and video retrieval.} \RR{}{Furthermore, in addition to the remote sensing community, we note that this dataset could also contribute to the computer vision community.}

\ifCLASSOPTIONcaptionsoff
  \newpage
\fi

\bibliographystyle{IEEEtran}
\bibliography{IEEEfull,reference}

\end{document}